\documentclass{article}
\usepackage{spconf,amsmath,graphicx}
\usepackage{amssymb}
\usepackage[hidelinks]{hyperref}
\usepackage{multirow}
\usepackage{makecell}
\usepackage{xcolor}
\definecolor{um2color}{HTML}{6A6A6A}
\definecolor{um4clapcolor}{HTML}{E69F00}
\definecolor{um4whispercolor}{HTML}{56B4E9}
\definecolor{mm6clapcolor}{HTML}{009E73}
\definecolor{mm6whispercolor}{HTML}{CC79A7}
\definecolor{mm63clapcolor}{HTML}{0072B2}
\definecolor{mm63whispercolor}{HTML}{D55E00}

\title{A Multimodal Approach to Device-Directed Speech Detection with Large Language Models}
\name{
\parbox{\linewidth} {
\centering
Dominik Wagner$^{1,*}$\thanks{$^{*}$Work done during an internship at Apple.}, Alexander Churchill$^{2}$, Siddharth Sigtia$^{2}$, Panayiotis Georgiou$^{2}$, Matt Mirsamadi$^{2}$,\\ Aarshee Mishra$^{2}$, Erik Marchi$^{2}$}
}
\address{$^{1}$Technische Hochschule Nürnberg, $^{2}$Apple}
\usepackage[margin=0.75in]{geometry} 
\makeatletter
\renewcommand\section{\@startsection {section}{1}{\z@}%
                                    {-2.3ex \@plus 0.0ex \@minus -.2ex}%
                                    {1.5ex \@plus.12ex}%
                                    {\normalfont\Large\bfseries}}
\renewcommand\subsection{\@startsection{subsection}{2}{\z@}%
                                      {-1.5ex \@plus 0.0ex \@minus -.2ex}%
                                      {0.75ex \@plus 0.12ex}%
                                      {\normalfont\large\bfseries}}
\makeatother                                      
\begin{document}
\ninept
\maketitle
\begin{abstract}
Interactions with virtual assistants typically start with a predefined trigger phrase followed by the user command.
To make interactions with the assistant more intuitive, we explore whether it is feasible to drop the requirement that users must begin each command with a trigger phrase.  
We explore this task in three ways: First, we train classifiers using only acoustic information obtained from the audio waveform. 
Second, we take the decoder outputs of an automatic speech recognition (ASR) system, such as 1-best hypotheses, as input features to a large language model (LLM).
Finally, we explore a multimodal system that combines acoustic and lexical features, as well as ASR decoder signals in an LLM.  
Using multimodal information yields relative equal-error-rate improvements over text-only and audio-only models of up to 39\% and 61\%. 
Increasing the size of the LLM and training with low-rank adaption leads to further relative EER reductions of up to 18\% on our dataset. 
\end{abstract}
\begin{keywords}
device-directed speech detection, large language model, multimodal, conditional generation
\end{keywords}
\vspace{-1mm}
\section{Introduction}
\label{sec:intro}
Speech based virtual assistants allow users to interact with phones, watches, mixed reality headsets, loudspeakers, and headphones via voice commands. 
Usually, a trigger phrase or the press of a button precedes the first command, to distinguish audio that is directed towards the device from background speech \cite{voicetrigger23}. 
The problem of detecting a trigger phrase is referred to as voice trigger detection \cite{sigtia18_interspeech,sigtia20vtd}, wake-word detection \cite{jose20_interspeech,ghosh22_interspeech}, or keyword spotting \cite{sainath15kws,michaely17kws,ng23kws}. 
For a more natural dialog flow, commands following the initial interaction should not need to include the trigger phrase.  
Device-directed speech detection addresses the problem of whether a virtual assistant was addressed or not, without a trigger cue preceding the voice command at all times \cite{shriberg12_interspeech,mallidi18_interspeech,garg22_interspeech}. 
This problem is significantly more challenging than voice trigger detection, since there might not be a leading trigger phrase that marks the beginning of a voice command. 
Many studies employ combinations of features such as acoustic, lexical, and even accelerometer signals, using a range of methods from late fusion to transformer-based neural networks for device-directedness detection \cite{shriberg12_interspeech,mallidi18_interspeech,gillespie20dd,buddi23multimodal,sato23multimodal,bekal22_interspeech}. 

Large language models (LLMs) have demonstrated state-of-the-art text comprehension abilities over a wide range of tasks \cite{radford2019gpt2,brown2020gpt3,openai2023gpt4}. 
This makes them interesting for device-directed speech decisions, where relevant information from all kinds of in-domain (e.g. voice commands) and out-of-domain signals (e.g. background speech, ambient sounds, appliances etc.) needs to be processed. 
Previous studies have extended LLMs with the ability to process non-text input modalities, such as audio and video data. 
The multimodal PaLM-E \cite{driess2023palme} combines a large pretrained vision transformer \cite{dehghani2023vit} with the PaLM LLM \cite{chowdhery2022palm} to perform robotic planning, as well as visual and language tasks. 
In \cite{fathullah2023prompting}, a LLaMA-7B \cite{touvron2023llama} LLM is finetuned to perform ASR by conditioning the model on a variable length sequence of audio representations. 
In \cite{gong2023ltu}, a pretrained audio spectrogram transformer \cite{gong21b_interspeech} provides audio representations to a pretrained LLaMA-7B LLM that is finetuned using low-rank adaptation (LoRA) \cite{hu2022lora} to perform a variety of tasks. 
ClipCap \cite{mokady2021clipcap} operates as an image captioning system by using representations from a pretrained CLIP \cite{radford2021clip} model as prefixes in a pretrained frozen GPT2 \cite{radford2019gpt2}. 
In \cite{kim2023prefix}, audio captioning is performed using a convolutional audio encoder that generates prefixes for a pretrained and frozen GPT2 model.  
In \cite{deshmukh2023pengi}, an audio encoder and a text encoder based on CLAP \cite{elizalde2022clap} provide representations that are used as prefixes to a pretrained frozen GPT2 model, which performs downstream tasks such as audio captioning, speech emotion recognition and question answering. 
Other recent studies have extended LLMs with the ability to process and understand audio inputs \cite{radhakrishnan2023whisperingllama,chu2023qwenaudio,tang2023salmonn}.

In this work, we explore a multimodal detection system to distinguish device-directed utterances from non-directed audio in interactions with a virtual assistant. 
Our objective is to determine whether a user addressed the assistant from the streaming audio captured by the device's microphone. 
The backbone is a pretrained LLM, which promises to offer advanced language understanding, multi-skill capabilities and ability to exploit turn taking as well as large context windows \cite{radford2019gpt2}. 

We are motivated to enhance the LLM with non-text signals by the observation that different modalities have disjoint weaknesses. 
Acoustic information alone is not reliable under background noise and overlapping speech, while lexical information suffers from errors due to ambiguous or poorly transcribed utterances.
Furthermore, this study is inspired by the recent successful efforts to equip LLMs with signals from non-lexical modalities and their ability to perform a wide range of new tasks. 
The present work uses multimodal information consisting of acoustic features obtained from a pretrained audio encoder, as well as 1-best hypotheses and utterance-level decoder signals, such as acoustic cost and graph cost from an ASR system. 
The acoustic features and decoder signals are represented as prefix tokens and then concatenated with the token embeddings of the 1-best hypotheses, to form the input to a pretrained LLM. 
The system is finetuned to generate device-directedness decisions by jointly learning from all modalities (cf. Figure \ref{fig:arch}). 

\vspace{-2mm}
\section{Data}
\vspace{-2mm}
\subsection{Training Data}
The training data comprised a balanced randomized and anonymized set of $\approx$40k directed utterances ($\approx$59 hours, mean duration 5.2$\pm$6.9 sec) and $\approx$40k non-directed utterances ($\approx$67 hours, mean duration 6.0$\pm$5.3 sec), similar to the set used in \cite{rudovic23sdsd} and \cite{dighe23a2i}. 
Approximately 21\% of the device-directed examples in the training set start with a trigger phrase. 
The remaining device-directed utterances represent triggerless interactions with a virtual assistant. 
\vspace{-3mm}
\subsubsection{Text-only Training 
Data}\label{sssec:add_text}
\vspace{-1mm}
An additional text corpus consisting of $\approx$3 million randomized and anonymized utterances of ASR-transcribed near-field speech signals recorded with devices such as smartphones is used as additional text-only information in our experiments. 
Similar data was used to finetune the text-based out-of-domain language detection component in \cite{voicetrigger23}. 
The corpus is split into $\approx$1.96M device-directed and $\approx$1.98M non-directed examples. 
Approximately 29\% of the device-directed examples start with a trigger phrase. 
A randomly sampled subset of these 500k text-only utterances was combined with the multimodal training set in some of our experiments, to simulate dropout of the audio modality \cite{neverova16moddrop}. 
\subsection{Evaluation Data}\label{ssec:eval_data}
For evaluation, we combine two randomized and anonymized in-house datasets with a total of $\approx$14k device-directed examples (mean duration 3.0$\pm$1.9 sec) and $\approx$23k non-directed examples (mean duration 3.7$\pm$3.6 sec). 
The total duration of the evaluation data is $\approx$35 hours. 
Approximately 12.3\% of the device-directed utterances start with a trigger phrase. 
The remaining device-directed utterances are triggerless interactions with a virtual assistant. 
\section{Feature Extraction}
\subsection{Text and ASR Features}\label{ssec:asr}
The text portion of our data was transcribed with an on-device joint CTC-attention based end-to-end speech recognition system \cite{kim17ctcatt}, representative of what a device-directedness detector would see at decision time. 
The ASR model is similar to the system used in \cite{bleeker23_interspeech}. 
It is trained on in-domain datasets and employs a Conformer \cite{gulati20_interspeech} as the acoustic encoder with a subword unit segmentation based on byte pair encodings (BPEs) \cite{sennrich16bpe}. 

Similar to \cite{shriberg12_interspeech,mallidi18_interspeech}, we extract 4 additional utterance-level signals that are generated by a decoder based on weighted finite-state transducers \cite{miao15eesen}. 
For the most likely hypothesis in the N-best set of hypotheses, we extract the average of the graph cost associated with each word in the hypothesis, the average of the acoustic cost, and the average of the word-level posterior confidence scores. 
The graph cost is the sum of LM cost, transition probabilities, and pronunciation cost \cite{povey12wfst}.
The acoustic cost is the negative log-likelihood of the BPE tokens from the decoder. 
Additionally, we include the average number of alternative word options for each word in the 1-best hypothesis. 
Finally, we apply min-max scaling along each signal dimension across the dataset to scale the feature values into the unit interval $\left[ 0, 1\right]$. 

\subsection{Acoustic Features}
Although, we use an ASR system to obtain text transcriptions, we decided against using acoustic representations from the same system in favor of more universal audio encoder models. 
The system described in Section \ref{ssec:asr} is trained on in-domain speech data, while we are processing all types of audio signals such as background speech and ambient noise in addition to the speech directed to an assistant. 
We assume that including unbiased acoustic representations can provide useful additional information that is less likely to contain the same mistakes as the lexical information. 

We compare two audio encoder backbones in our experiments. 
The first model is the medium-sized (769M parameters) version of Whisper \cite{radford2022robust}, an encoder-decoder transformer \cite{vaswani17attention} for multilingual speech recognition trained on 680k hours of audio paired with transcripts.
Whisper is well-suited for our task, since it has seen a wide range of acoustic conditions and is expected to generalize well across domains and languages. 
Whisper operates on 80-dimensional log-magnitude Mel spectrogram features with a window length of 25ms and a stride of 10ms. 

The second model is CLAP \cite{elizalde2022clap} (Contrastive Language-Audio Pretraining), which is trained to represent text descriptions and audio in the same latent space by using two encoders and contrastive learning. 
The model is trained on audio captioning and sound event classification datasets.  
CLAP has approximately 153 million trainable parameters and operates on 64-dimensional log-magnitude Mel spectrograms with a window length of 21ms and a stride of 10ms.


\section{Method}
\begin{figure}[t]
\centering
\includegraphics[width=0.5\textwidth]{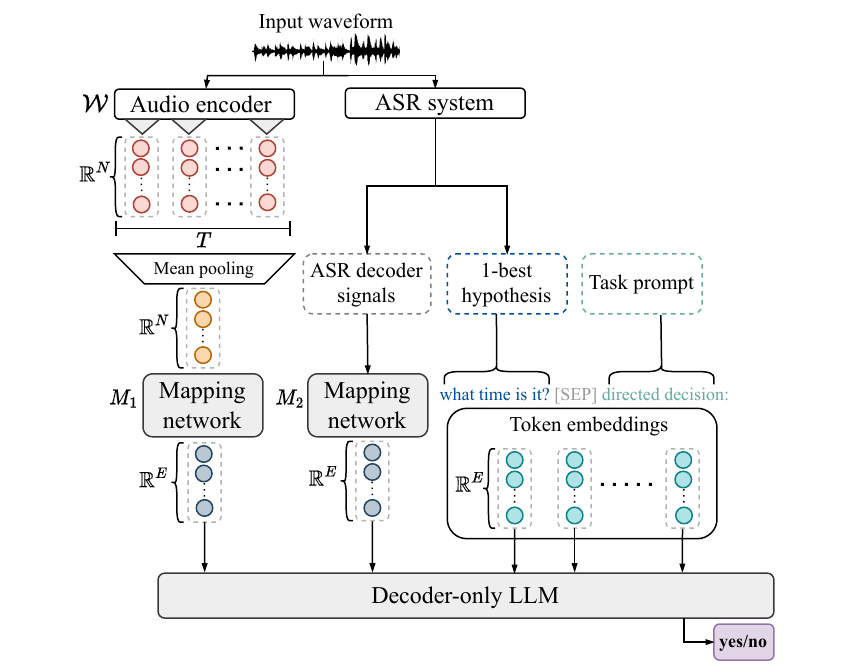}
\vspace{-8mm}
\caption{Architecture of the multimodal system. The weights of the grey-shaded components ($M_1$, $M_2$ and LLM) are trained, all other components remain frozen. The unimodal baselines differ from the multimodal system as follows: In the \textit{text-only} variant, the mapping networks $M_1$ and $M_2$ are removed and the only input features are the 1-best hypotheses of the ASR system. In the \textit{audio-only} variant, the decoder signals including $M_2$ and the 1-best hypotheses are removed. The \textit{DS-only} system relies only on the decoder signal input, which is transformed via $M_2$, i.e., $M_1$ and the 1-best hypotheses are removed from the overall system.}
\label{fig:arch}
\end{figure}
The architecture of our multimodal model is depicted in Figure \ref{fig:arch}. 
In the interest of brevity, we only describe the full multimodal version of our system. 
The unimodal versions used as baselines in Section \ref{sec:exp} are implemented by keeping only the necessary components for one modality (e.g. text) and by removing all components related to other modalities. 
The full multimodal model has three main components: (1) an audio encoder that extracts a sequence of latent representations in $\mathbb{R}^{N}$ from the input audio, (2) two mapping networks $M_1$ and $M_2$, that map the extracted audio features and the utterance-level ASR decoder signals into the token embedding space, and (3) a decoder-only LLM that generates text conditioned on the prefixes, the 1-best ASR hypotheses, and a text prompt. 

Given a training dataset with $L$ examples of audio, decoder signals, and 1-best hypotheses $\{(x^{(i)}, d^{(i)}, t^{(i)})\}_{i=1}^L$, the goal is the generation of a device-directedness decision (i.e.,  directed to a device ``yes'' or ``no'') for an unseen set of features. 
The 1-best hypothesis of the $(i)$-th example is represented by a sequence of tokens $t^{(i)} = ( t^{(i)}_1, \ldots, t^{(i)}_l )$ padded to a maximum length $l$.  

To extract representations from the $(i)$-th audio input $x^{(i)}$, we use the audio encoder $\mathcal{W}$ (either Whisper or CLAP), which yields a sequence of embeddings in $\mathbb{R}^{T \times N}$. 
We aggregate these embeddings along the time dimension $T$ using mean pooling, which outputs a single vector in $\mathbb{R}^{N}$ per utterance. 
Then, we employ a small feedforward network $M_1$ to map the aggregated embedding to a vector in the prefix token space of the LLM: 
\begin{equation}
    a^{(i)} = M_1 \left(\frac{1}{T} \sum_{t=1} ^T h_t \right), \; \mathcal{W} \left(\mathcal{F} \left( x^{(i)} \right) \right)=[h_1 \cdots h_T]^\top,
\end{equation}
where $\mathcal{F}$ is the transformation of the input waveform $x^{(i)}$ into log-magnitude Mel spectrogram features. 
Each representation $a^{(i)}$ has the same dimensionality $E$ as the token embeddings. 

We use a second feedforward network $M_2$ to extract a latent representation for the decoder signals $d^{(i)}$, which yields a prefix $b^{(i)}$ in $\mathbb{R}^E$.  
We then concatenate the audio prefix $a^{(i)}$ and the decoder signal prefix $b^{(i)}$ to the token embeddings of the corresponding 1-best hypothesis $t^{(i)}$. 
The language model is presented with the concatenated input features. 
The training objective is the prediction of directedness tokens conditioned on the audio prefix, the decoder signal prefix, and the 1-best hypothesis tokens in an autoregressive fashion. 
We train the model using cross entropy loss:
\begin{equation}
    \mathcal{L}_\theta =-\sum_{i=1}^L \sum_{j=1}^l \log p_\theta\left(t_j^{(i)} \mid a^{(i)}, b^{(i)}, t_1^{(i)}, \ldots, t_{j-1}^{(i)} \right),
\end{equation}
where $\theta$ are the trainable parameters of the model. 
During inference, we consider the score $p_\theta\left(Y=yes \mid c \right)$ to make the device-directedness decision, where $Y$ is a discrete random variable that can realize one of $m$ tokens  $y_1, ... , y_m$ from the vocabulary $\mathcal{V}$ and $p_\theta\left(Y=yes \mid c \right) + p_\theta\left(Y=no \mid c \right) \simeq 1$. 
The context $c$ is given by the multimodal features, i.e., $c = (a^{(i)}, b^{(i)}, t_1^{(i)}, \ldots, t_{j-1}^{(i)})$. 
\subsection{Large Language Models}
We explore the 124.4M and 1.5B parameter versions of GPT2 in our experiments. 
Several studies show that GPT2 can be adapted to new text generation tasks by providing learnable preﬁxes \cite{mokady2021clipcap,kim2023prefix,deshmukh2023pengi}. 
Furthermore, decoder-only LLMs have demonstrated stronger capabilities \cite{brown2020gpt3} than encoder-only and encoder-decoder systems such as BERT \cite{devlin-etal-2019-bert} and T5 \cite{raffel20t5} on a wide range of tasks. 
We choose the small version of GPT2 for the main part of our experiments, since it is relatively lightweight and can potentially run on devices such as smartphones. 
Since our primary concern is not to maintain the expressivity of the LLM for text generation, but to receive accurate device-directedness decisions, we finetune the GPT2 weights during training along with the mapping networks. 
Nevertheless, we also analyze the impact of increased model size and parameter-efficient finetuning with LoRA \cite{hu2022lora}, since this approach yielded promising results in our preliminary study \cite{wagner2023multimodal}. 
\subsection{Mapping Networks}
During training, we face the challenge of translating between the latent space of the audio encoder and the lexical embedding space of the LLM. 
Although both models generate rich and diverse representations, their latent spaces are independent. 
The mapping networks $M_1$ and $M_2$ are designed to bridge the gap between the two latent spaces and to meaningfully connect audio features and decoder signals with GPT2. 
The outputs of the mapping networks are used as prefixes to the text tokens. 
Each audio feature is translated to $\mathbb{R}^{E}$ sized prefixes, where $E$ is the latent dimension of GPT2. 
Both mapping networks share the same architecture. 
They consist of 1 hidden linear layer with 384 units, hyperbolic tangent activation and are trained with a dropout probability of 10\%. 

\subsection{Training Details}
We use the same training procedure for the unimodal and the multimodal versions of our system. 
We extract 1024-dimensional representations at the last encoder layer of the pretrained Whisper model and 512-dimensional representations at the audio projection layer of the pretrained CLAP model. 
Our system is trained for 60 epochs with an effective batch size of 256. 
For optimization, we use AdamW \cite{loshchilov2018decoupled} with an initial learning rate of $1 \times 10^{-4}$, a linear schedule and a warm-up phase of 10\% of total training steps. 
The text token sequences are padded to $l=256$. 

\subsection{Unimodal Baselines}
We implement the following unimodal baseline systems: 
First, we train linear audio-only classifiers on top of Whisper and CLAP. 
Their classification heads consist of a linear layer with 256 hidden units, which receives representations obtained from the audio encoder model (i.e. Whisper or CLAP), followed by mean pooling across the time dimension and a final linear layer. 
The classifiers are trained for 10 epochs with an effective batch size of 32 using cross entropy loss. 
We use the AdamW optimizer with an initial learning rate of $2 \times 10^{-5}$, and a warm-up phase of 10\% of the total training steps with a linear schedule. 
During training, all parameters of the respective audio encoder remain frozen. 
Second, we train three unimodal versions of our framework (cf. Figure \ref{fig:arch}) by providing either text, decoder signals, or audio representations as the only input source to the LLM. 
We also explored zero-shot learning on our task. 
However, zero-shot accuracy was close to random, which is expected given the GPT2 models have not been instruction-finetuned.
\vspace{-2mm}
\section{Experiments}\label{sec:exp}
\begin{table}[t]
\setlength{\tabcolsep}{1pt}
    \caption{Comparison of equal-error-rates (EERs) on the evaluation set. ``CLF'' is the frozen audio encoder with a linear classification head, ``UM'' refers to unimodal experiments, and ``MM'' refers to multimodal experiments. The column ``Modality'' indicates the modalities used in the experiment (text = 1-best hypothesis $t^i$, audio = audio representations $a^i$, DS = decoder signals $b^i$). ``Add. Text'' shows the number of examples from the additional text corpus. ``Trainable Params'' is the total number of trainable parameters, i.e., adapter and mapping networks in millions. When $M_1$ was involved, we used the trainable parameters of the larger mapping network (i.e., the one receiving Whisper features with $\approx$0.7M parameters, as opposed to $\approx$0.5M for CLAP features) to compute the total number of parameters. Colors correspond to the DET curves in Figure \ref{fig:det}.} 
  \centering \def\arraystretch{1.5} \small
\begin{tabular}{ |c|c|c|c|c|c|c| } 
 \hline
  \textbf{Exp.}               & \textbf{LLM}       & \textbf{Modality}        & \makecell{\textbf{Add.} \\ \textbf{Text}}     & \makecell{ \textbf{EER} \\ \textbf{Whisper} } & \makecell{ \textbf{EER} \\ \textbf{CLAP}}  & \makecell{ \textbf{Trainable} \\ \textbf{Params}} \\
  \hline
   \multicolumn{7}{|c|}{\textbf{Unimodal baselines}} \\
    \hline
  CLF    & --   & audio-only               &  --  & 16.70\% & 23.47\% & 0.3M \\
 UM1                      &  GPT2       & text-only   &  --       &    12.70\% & 12.70\% & 124.4M \\
 UM2                     &  GPT2      & text-only    &  500k   &       \textcolor{um2color}{12.15\%} & \textcolor{um2color}{12.15\%} & 124.4M \\
 UM3                     &  GPT2     & DS-only       &  --     & 28.09\% & 28.09\%  & 125.1M      \\
 UM4                      &  GPT2     & audio-only    &  --     & \textcolor{um4whispercolor}{10.98\%}      & \textcolor{um4clapcolor}{19.13\%} & 125.1M \\
 \hline
 \multicolumn{7}{|c|}{\textbf{Multimodal experiments -- Full finetuning}} \\
  \hline
    MM1                      &  GPT2 & text+DS               &  --          &  10.54\%  & 10.54\% & 125.1M \\
 MM2                       &  GPT2 & audio+DS               &  --         &  8.81\%  & 16.05\% & 125.8M\\
 MM3                    &  GPT2 & text+audio               &  --       &  8.61\%  & 9.34\% & 125.1M\\
  MM4                   &  GPT2 & text+audio              &  500k      & 8.59\%  & 9.00\% & 125.1M\\
 MM5                   &   GPT2 & text+audio+DS            & --       & 8.07\% & 7.77\% & 125.8M\\
  MM6                 &   GPT2 & text+audio+DS           & 500k      & \textcolor{mm6whispercolor}{\textbf{7.95\%}} & \textcolor{mm6clapcolor}{\textbf{7.45\%}} & 125.8M \\
  \hline
\end{tabular}
  \label{tab:exp}
\end{table}
Table \ref{tab:exp} shows the equal-error-rates (EERs) on the evaluation set. 
EER is the location on a detection error trade-off (DET) curve where the false acceptance rate (FAR) and the false rejection rate (FRR) are equal, i.e., both error types are considered equally important \cite{martin97b_eurospeech}. 
The unimodal baselines are the frozen Whisper and CLAP models with a linear classification head (CLF), as well as text-only, audio-only, and decoder signal (DS) only versions of our proposed system (UM1-4). 
Including additional in-domain text data improves EER by 4.3\% relative to the text-only model without additional text data (UM1 vs. UM2). 
The DS in experiment UM3 provide a weak signal on their own ($EER=28.1\%$). 
Using our system with only the audio prefix and no text information (UM4) improves EER relative to CLF by 34.3\% and 18.5\%, highlighting the effectiveness of replacing the simple linear classification head with the more complex LLM. 

The best system configuration (MM6) leverages 500k examples of additional text data (cf. Section \ref{sssec:add_text}) and combines text, audio, as well as decoder signals from the ASR system. 
Experiment MM6 yields an EER of 7.95\% with the Whisper audio encoder and an EER of 7.45\% with the CLAP backbone, which translates to relative improvements of 27.6\% and 61.1\% over the corresponding audio-only models (UM4). 
The relative EER improvements over the best performing text-only model (UM2) are 34.6\% and 38.7\%. 

Using CLAP as the audio encoder performs consistently worse than Whisper, as long as not all available modalities are used (MM1-4). 
However, when all three signals are combined (MM5 and MM6), the features obtained with CLAP are more effective than those of Whisper. 


\begin{table}[t]
\setlength{\tabcolsep}{2pt}
    \caption{Comparison of equal-error-rates (EERs) for the best models in Table \ref{tab:exp} (MM6) with LoRA and increased LLM size. The hyperparameter $r$ is the rank of the matrices used for adaptation, and $\alpha$ is a scaling factor to adjust the magnitude of the adaptation.}
  \centering \def\arraystretch{1.5} \small
\begin{tabular}{ |c|c|c|c|c|c| } 
 \hline
  \textbf{Exp.} & \textbf{LLM} & \makecell{ \textbf{LoRA} \\ \textbf{Configuration} } & \makecell{ \textbf{EER} \\ \textbf{Whisper} } & \makecell{ \textbf{EER} \\ \textbf{CLAP} } & \makecell{ \textbf{Trainable} \\ \textbf{Params} }\\
  \hline
  MM6.1                     &   GPT2        & {\footnotesize $r=64$, $\alpha = 16$} & 9.13\% & 9.32\% & 3.7M \\
  MM6.2                     &   GPT2 \scriptsize{1.5B}   & {\footnotesize $r=8$, $\alpha = 32$} & 6.75\% & 8.51\% & 5.2M \\
  MM6.3                     &   GPT2 \scriptsize{1.5B}   & {\footnotesize $r=64$, $\alpha = 16$} & \textcolor{mm63whispercolor}{\textbf{6.53\%}} & \textcolor{mm63clapcolor}{\textbf{8.41\%}} & 22.5M \\
  \hline
\end{tabular}
  \label{tab:abl}
\end{table}
One disadvantage of the method used in Table \ref{tab:exp} is that by finetuning the LLM directly, we may lose its ability to perform a wide range of tasks and full finetuning becomes less feasible with increasing model size. 
In a more practical scenario, only a single LLM that can not be directly finetuned might be available. 
Therefore, the experiments in Table \ref{tab:abl} show the EER with LoRA \cite{hu2022lora} adapters attached to the transformer layers of the LLM. 
LoRA adapters are small trainable matrices that are included into each layer of the transformer architecture and optimized instead of the underlying LLM weights, thereby reducing the number of trainable parameters for downstream tasks. 
We find that applying LoRA to the small version of GPT2 (MM6.1) is less effective than full finetuning (MM6 in Table \ref{tab:exp}). 
Using the GPT2 1.5B model yields EERs of 6.53\% and 8.41\% (MM6.3), which is a considerable improvement over experiment MM6 in Table \ref{tab:exp} with the Whisper backbone, but an EER increase with the CLAP backbone. 

\begin{figure}[t]
\centering
\includegraphics[width=0.5\textwidth]{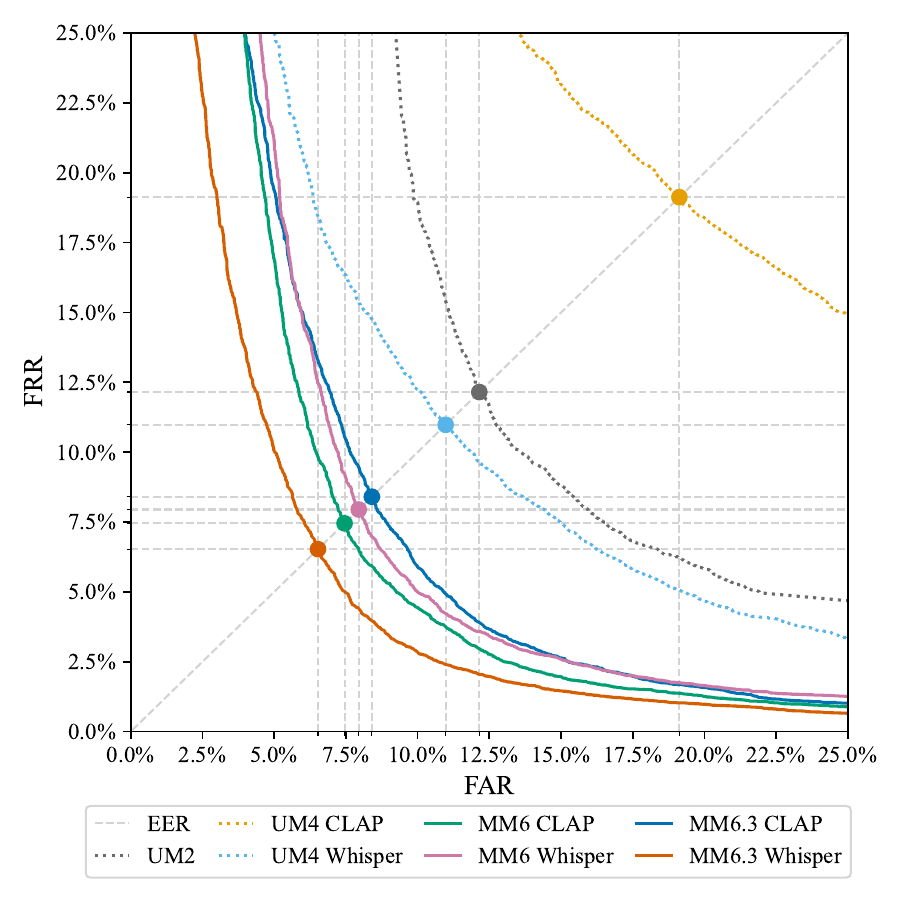}
\vspace{-10mm}
\caption{DET curves for a selection of experiments from Table \ref{tab:exp} and Table \ref{tab:abl}. The false accept rate (FAR) represents non-directed utterances that were falsely classified as directed utterances and the false reject rate (FRR) represents directed utterances that were falsely classified as non-directed utterances. 
Dotted lines show unimodal baselines (UM2 and UM4) and solid lines show multimodal experiments (MM6 and MM6.3). 
The points on each curve indicate the EER of the respective experiment. 
}
\label{fig:det}
\vspace{-3mm}
\end{figure}

The detection error trade-off (DET) curves for a selection of experiments are depicted in Figure \ref{fig:det}. 
We show an operating region for false rejects and false accepts of $\leq25\%$. 
The multimodal models (solid lines) are more effective than the unimodal models (dotted lines), irrespective of the audio encoder backbone used. 
The best system (MM6.3) shows lower FARs and FRRs than the other experiments across the entire operating region. 
\subsection{Discussion}
We find that CLAP representations on their own provide a weaker signal than Whisper representations (cf. UM4 in Table \ref{tab:exp}). 
Mainly due to its training on ASR, we believe Whisper is more capable of capturing information that is also included in text (i.e., 1-best hypotheses) and in ASR decoder signals. 
Consequently, the relative EER improvements from including additional modalities are smaller than with the CLAP backbone. 
Once text and ASR decoder features are added, they seem to compensate for CLAP shortcomings and in fact using the CLAP backbone, which is trained on acoustic scene characteristics, outperforms Whisper features in this case (cf. MM6 in Table \ref{tab:exp}). 
However, the EER also depends on the ability to perform full finetuning, as well as the size of the LLM. 
Using the Whisper backbone with LoRA and a larger LLM yields improvements over full finetuning, whereas the same experiment with CLAP does not. 
Therefore, using acoustic information from the Whisper model seems to be a more robust choice in parameter-efficient finetuning scenarios. 

\vspace{-2mm}
\section{Conclusions}
We described a multimodal model to distinguish device-directed utterances from background speech in the context of interactions with a virtual assistant. 
We provided audio representations and ASR decoder signals as additional inputs to an LLM and showed that the system is able to effectively combine decoder signals with audio and lexical information. 
The lowest overall EER improved upon the best unimodal system by 40\%. 
This was achieved by using all three modalities with the largest available version of GPT2 in combination with parameter-efficient finetuning and Whisper as the audio encoder backbone. 

Future work will focus on enhancing our approach with additional tasks that can be useful during interactions with virtual assistants, such as audio captioning and acoustic scene classification. 
Additionally, we will explore longer contexts, e.g., by including features from previous interactions.  
\vspace{-1mm}
\section{Acknowledgements}
We would like to thank John Bridle, Pranay Dighe, Oggi Rudovic, Ahmed Tewfik,  Barry Theobald and Sachin Kajarekar for their support and their comprehensive feedback on the paper. 
We also thank Seanie Lee for the moral support and numerous helpful discussions.  
\newpage
\bibliographystyle{IEEEbib}
\footnotesize{
    \bibliography{refs}
}

\end{document}